\documentclass[conference]{IEEEtran}
\IEEEoverridecommandlockouts
\usepackage{cite}
\usepackage{amsmath,amssymb,amsfonts}
\usepackage{algorithmic}
\usepackage{graphicx}
\usepackage{textcomp}
\usepackage{xcolor}
\usepackage{subfigure}
\usepackage{array}
\usepackage{float}
\def\BibTeX{{\rm B\kern-.05em{\sc i\kern-.025em b}\kern-.08em
    T\kern-.1667em\lower.7ex\hbox{E}\kern-.125emX}}
\begin{document}

\title{Sequential Spatial Network for Collision Avoidance in Autonomous Driving
\thanks{This research is supported by Sino German fond (5091331).}
}

\author{\IEEEauthorblockN{1\textsuperscript{st} Haichuan Li}
\IEEEauthorblockA{\textit{Chair of Robotics, Artificial } \\ 
\textit{Intelligence and Real-Time Systems} \\
\textit{Technical University of Munich}\\
Garching, Germany \\
haichuan.li@tum.de}
\and
\IEEEauthorblockN{2\textsuperscript{nd} Liguo Zhou}
\IEEEauthorblockA{\textit{Chair of Robotics, Artificial } \\ 
\textit{Intelligence and Real-Time Systems} \\
\textit{Technical University of Munich}\\
Garching, Germany \\
liguo.zhou@tum.de}
\and
\IEEEauthorblockN{3\textsuperscript{rd} Zhenshan Bing}
\IEEEauthorblockA{\textit{Chair of Robotics, Artificial } \\ 
\textit{Intelligence and Real-Time Systems} \\
\textit{Technical University of Munich}\\
Munich, Germany \\
zhenshan.bing@tum.de}
\and
\IEEEauthorblockN{4\textsuperscript{th} Marzana Khatun}
\IEEEauthorblockA{\textit{Institute for Driver Assistance and} \\
\textit{ Connected Mobility} \\
\textit{Hochschule Kempten}\\
Kempten, Germany \\
marzana.khatun@hs-kempten.de}
\and
\IEEEauthorblockN{5\textsuperscript{th} Rolf Jung}
\IEEEauthorblockA{\textit{Institute for Driver Assistance and} \\
\textit{ Connected Mobility} \\
\textit{Hochschule Kempten}\\
Kempten, Germany \\
rolf.jung@hs-kempten.de}
\and
\IEEEauthorblockN{6\textsuperscript{th} Alois Knoll}
\IEEEauthorblockA{\textit{Chair of Robotics, Artificial Intelligence} \\ 
\textit{ and Real-Time Systems} \\
\textit{Technical University of Munich}\\
Garching, Germany \\
knoll@in.tum.de}
}

\maketitle

\begin{abstract}

Several autonomous driving strategies have been applied to autonomous vehicles, especially in the collision avoidance area. The purpose of collision avoidance is achieved by adjusting the trajectory of autonomous vehicles (AV) to avoid intersection or overlap with the trajectory of surrounding vehicles. A large number of sophisticated vision algorithms have been designed for target inspection, classification, and other tasks, such as ResNet, YOLO, etc., which have achieved excellent performance in vision tasks because of their ability to accurately and quickly capture regional features. However, due to the variability of different tasks, the above models achieve good performance in capturing small regions but are still insufficient in correlating the regional features of the input image with each other. In this paper, we aim to solve this problem and develop an algorithm that takes into account the advantages of CNN in capturing regional features while establishing feature correlation between regions using variants of attention. Finally, our model achieves better performance in the test set of L5Kit compared to the other vision models. The average number of collisions is 19.4 per 10000 frames of driving distance, which greatly improves the success rate of collision avoidance.

\end{abstract}

\begin{IEEEkeywords}
Collision Avoidance, Computer Vision, Autonomous Driving, Trajectory Prediction, Sequential Spatial
\end{IEEEkeywords}

\section{Introduction}

In the past many years, researchers have focused on how to turn vehicles from assisted driving to more intelligent autonomous driving. Due to the iteration of intelligent hardware and the improvement of chip computing power, a large amount of data collected by sensors can be quickly converted and fed into models to make decisions. In the driving process, the safety factor is the first consideration for users and researchers. Therefore, how AV should avoid collisions has become a top priority.  Concepts such as probabilistic methods (eg.: Markov chains~\cite{chung1967markov} and Monte 
\begin{figure}[t]
    \centering
    \subfigure[Front Collision]{\includegraphics[width=0.48\textwidth]{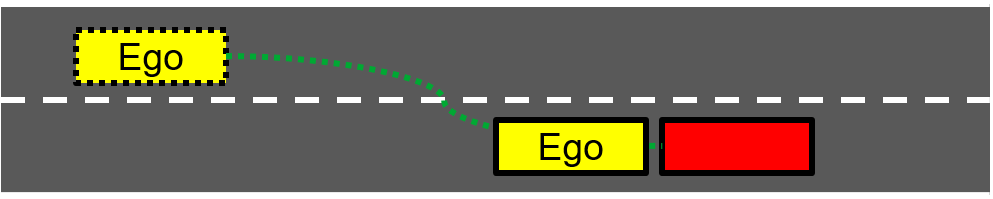}}
    \subfigure[Side Collision]{\includegraphics[width=0.48\textwidth]{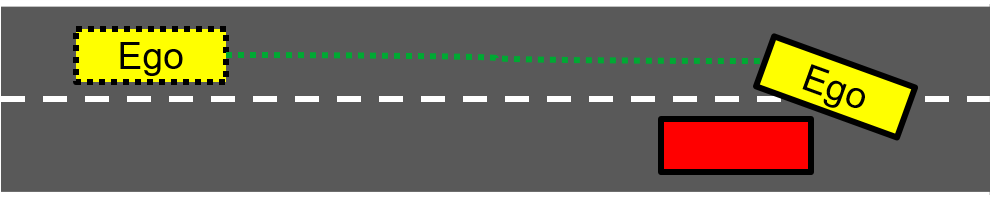}}
    \subfigure[Rear Collision]{\includegraphics[width=0.48\textwidth]{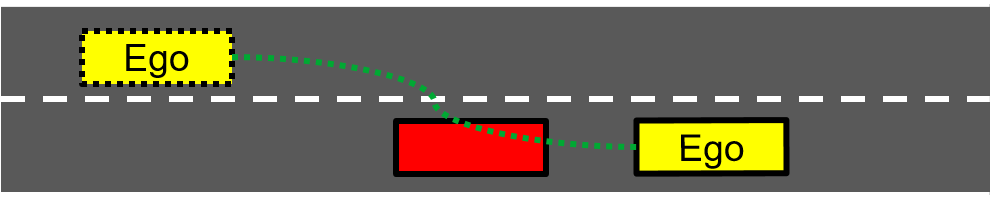}}
    \caption{Three different situations of collisions. The front collision case is shown in (a). The side collision case is shown in (b). The rear collision case is shown in (c). Our purpose is to let autonomous driving vehicles avoid collisions such as these.}
    \label{fig1}
\end{figure}
Carlo~\cite{metropolis1949monte}), safety distance-based control methods~\cite{hamid2018review}, and trajectory prediction methods~\cite{hamid2018review} have been designed in recent years to cope with complex traffic conditions. In terms of vision, CNN~\cite{lecun1998gradient} has made outstanding contributions and has been applied to a large number of road condition inspection tasks due to its excellent regional feature extraction capabilities. The local feature information obtained by CNN will be used for obstacle detection. Secondly, because the motion trajectory is planned for AV, the relationship between each local feature of the image obtained by CNN needs to be established. Some strategies are based on CNN plus RNN~\cite{rumelhart1985learning} so that they can deal with sequential graphs as input, eg.: STDN~\cite{yao2018modeling}.

Although the above strategies have performed well in a large number of vision tasks, their performances are still far inferior to similar-sized convolutional neural networks counterparts, such as EfficientNets~\cite{tan2019efficientnet} and RepVGG~\cite{ding2021repvgg}. We believe this is due to the following aspects. First, the huge differences between the sequential tasks of NLP and the image tasks of CV are ignored. For example, when the local feature information acquired in a two-dimensional image is compressed into one-dimensional time series information, how to achieve accurate mapping becomes a difficult problem. Second, it is difficult to keep the original information of inputs since after RNN layers, we need to recover the dimension from one to three. Besides, due to the several transformations between different dimensions, that process becomes even harder, especially since our input size is 224×224×5. Third, the computational and memory requirement of switching between layers are extremely heavy tasks, which also becomes a tricky point for the algorithm to run. Higher hardware requirements as well as more running time arise when running the attention part.

In this paper, we propose a new network structure based on CNN and attention to vision tasks in autonomous driving. The new network structure overcomes these problems by using Sequential Spatial Network (SSN) blocks. As shown in Fig.~\ref{fig3}, input images first go through the convolution stem for fine-grained feature extraction, and are then fed into a stack of SSN blocks for further processing. The Upsampling Convolutional Decreasing (UCD) blocks are introduced for the purpose of local information enhancement by deep convolution, and in SSN block of features generated in the first stage can be less loss of image resolution, which is crucial for the subsequent trajectory adjustment task. 

In addition, we adopt a staged architecture design using five convolutional layers with different kernel sizes and steps gradually decreasing the resolution (sequence length) and flexibly increasing the dimensionality. Such a design helps to extract local features of different scales and, since the first stage retains high resolution, our design can effectively reduce the resolution of the output information in the first layer at each convolutional layer, thus reducing the computational effort of subsequent layers. The Reinforcement Region Unit (RRU) and the Fast MultiHead Self-Attention (FMHSA) in the SSN block can help obtain global and local structural information within the intermediate features and improve the normalization capability of the network. Finally, average pooling is used to obtain better trajectory tuning. 

Extensive experiments on the lykit dataset demonstrate the superiority of our SSN network in terms of accuracy. In addition to image classification, SSN block can be easily transferred to other vision tasks and serve as a versatile backbone.

\section{Related Works}

Over the past few decades, autonomous driving has flourished in the wave of deep learning, where a large number of solution strategies are based on computer vision, using images as the primary input. The prevailing visual neural networks are typically built on top of a basic block in which a series of convolutional layers are stacked sequentially to capture local information in intermediate features. However, the limited receptive field of the small convolution kernel makes it difficult to obtain global information, which hinders the high performance of the network on highly feature-dependent tasks (such as trajectory prediction and planning). In view of this dilemma, many researchers have begun to deeply study self-attention-based~\cite{vaswani2017attention} networks with the ability to capture long-distance information. Here, we briefly review traditional CNNs and recently proposed visual networks. Convolutional neural network. The first standard CNN was proposed by LeCun~\cite{lecun1989backpropagation} et al. was used for handwritten character recognition. Based on this foundation, a large number of visual models have achieved cross-generational success in a variety of tasks with images as the main input. Google Inception Net ~\cite{szegedy2015going} and DenseNet ~\cite{huang2017densely} showed that deep neural networks consisting of convolutional and pooling layers can yield adequate results in recognition. SENet~\cite{hu2018squeeze} and MobileNetV3 ~\cite{howard2019searching} demonstrate the effectiveness of multiple paths within a basic block.

ResNet~\cite{he2016deep} is a classic structure that has a better generalization ability by adding shortcut connections to the underlying network. To alleviate the limited acceptance domain in previous studies, some studies used the attention mechanism as an operator for adapting patterns.

\begin{figure*}[!ht]
    \centering
    \subfigure[ResNet-50]{
    \includegraphics[height=8cm]{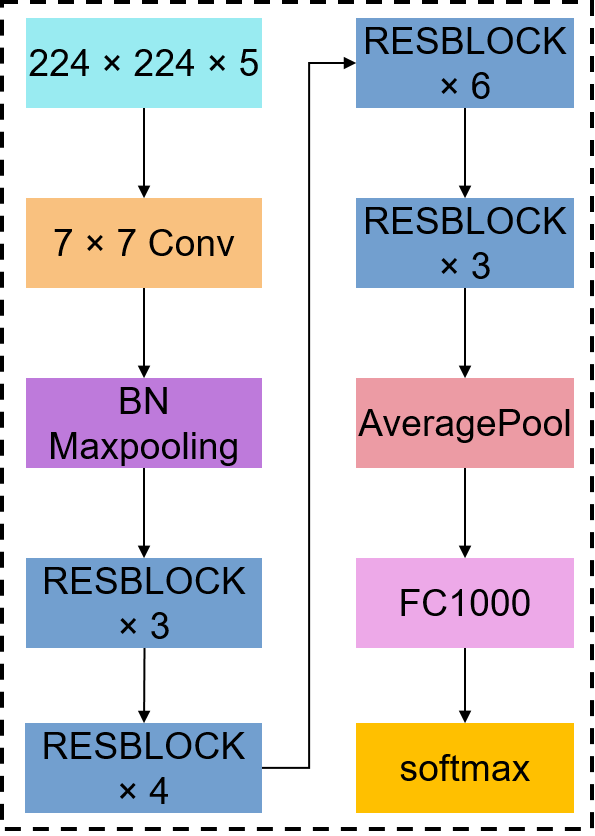}
    }
    \subfigure[RepVGG]{
    \includegraphics[height=8cm]{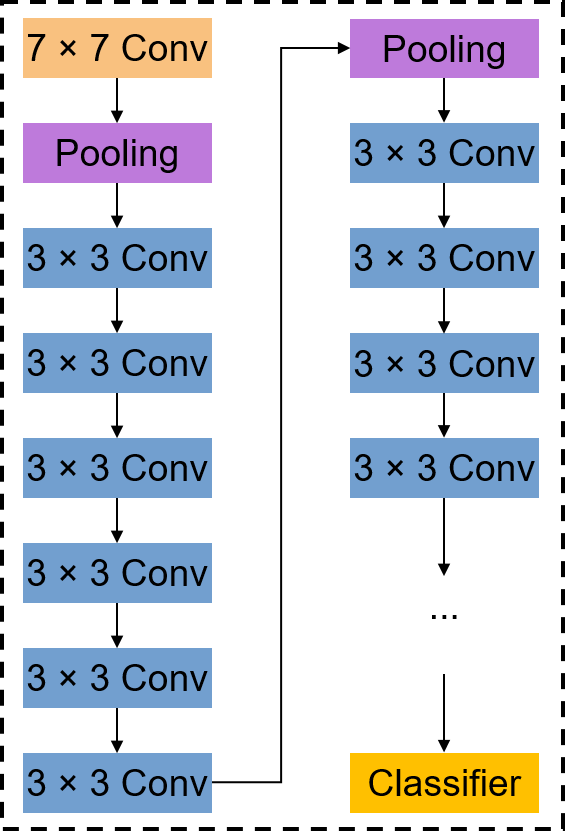}
    }
    \subfigure[ViT]{
    \includegraphics[height=8cm]{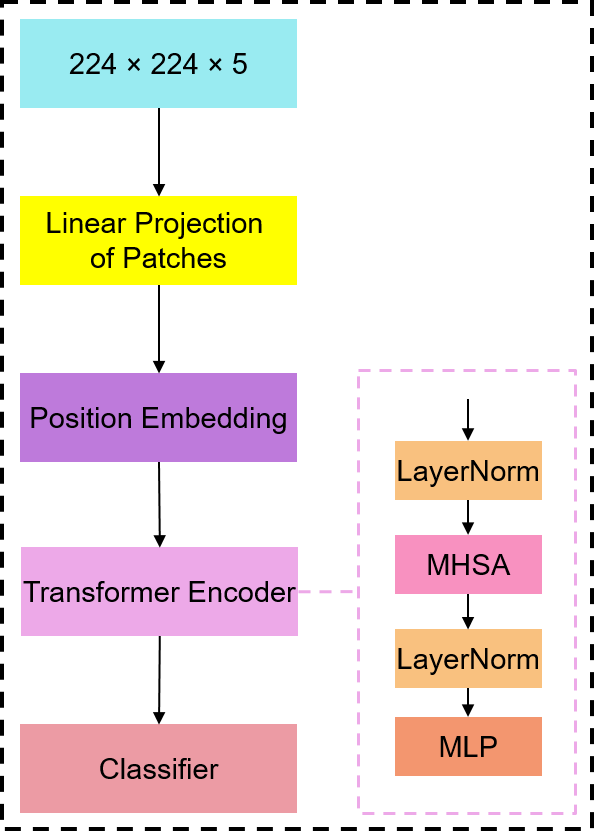}
    }
    \caption{Three popular network structures in vision areas. The structure of ResNet-50 is shown in (a). The structure of RepVGG is shown in (b). The structure of ViT is shown in (c).}
    \label{fig2}
\end{figure*}

\section{Approach}

\subsection{Network Structure Overview}

\begin{figure*}[h]
    \centering
    \includegraphics[width=\textwidth]{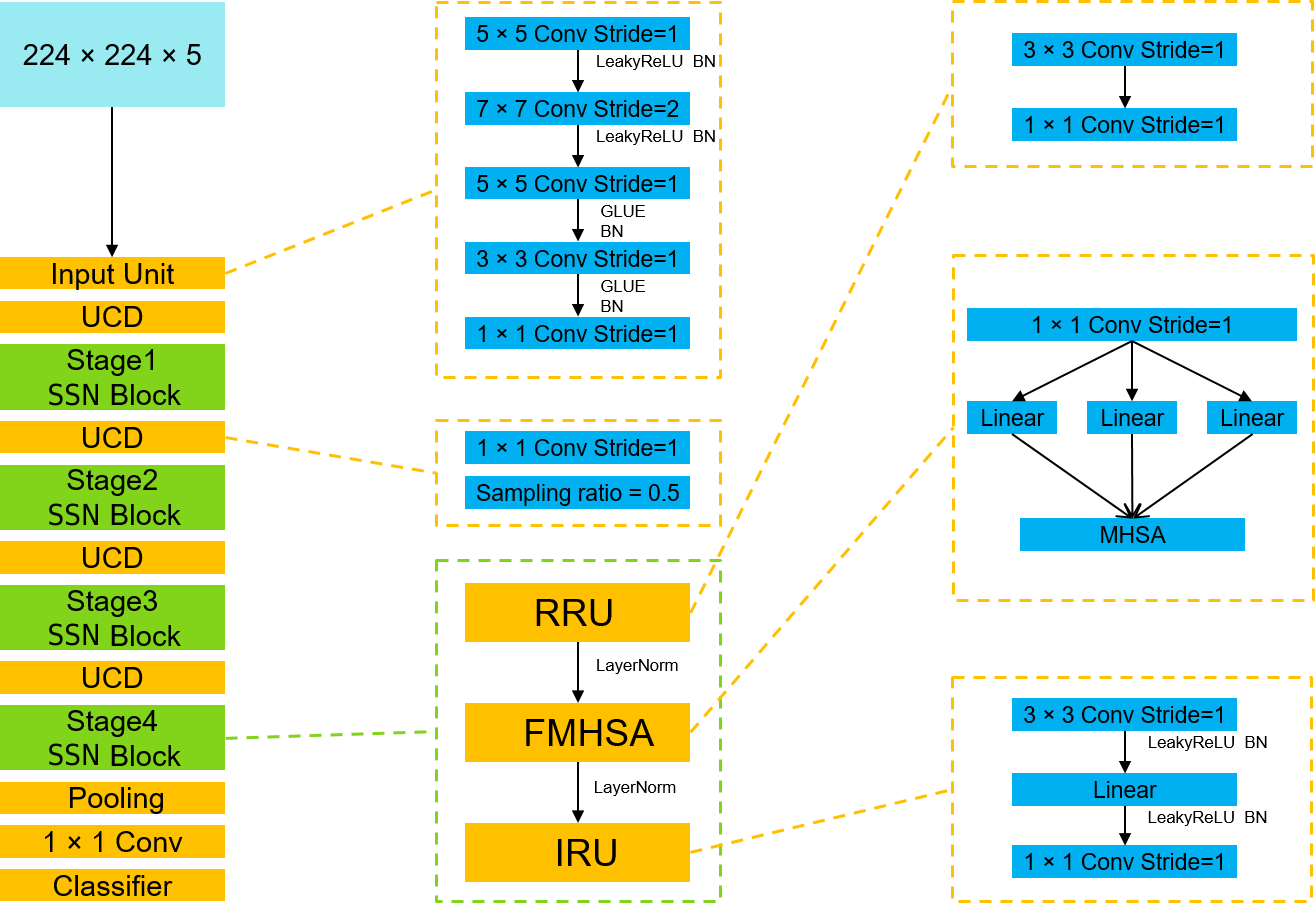}
    \caption{The overview of SSN network structure. RRU is Reinforcement Region Unit. FMHSA is Fast Multi-Head Self-Attention module. IRU is Information Refinement Unit.}
    \label{fig3}
\end{figure*}

Our strategy is to take advantage of both CNN and attention by building a hybrid network. An overview of ResNet-50~\cite{he2016deep}, RepVGG~\cite{ding2021repvgg}, ViT~\cite{dosovitskiy2020image} and our network are shown in Fig.~\ref{fig2} and Fig.~\ref{fig3}. 

Resnet-50 consists of five stages, stage0 consists of a convolutional layer, a batch normalization layer and a maxpooling layer. stage1, stage2, stage3, and stage4 consist of bottleneck blocks, and the output is fed to a fully connected layer for classification. The advantage of this design is that resnet50 can efficiently handle the classification problem between different images. However, since the stage0 only uses one convolutional layer with a large convolutional kernel, the large field of view can quickly complete the initial processing of the input image, but the local information capture of the image is slightly insufficient. For this reason, we use the main input block to deal with this limitation. The main input block is composed of five different kernel sizes and steps convolutional layers, so that the purpose of stepping down the convolution kernel is to extract the input information quickly when the output size is large, and then use the small convolution kernel to extract the local information after the input size becomes smaller. 

After taking into account the fast processing and local information processing of the main input block, the input information is transferred to the subsequent blocks for subsequent processing. Furthermore, between each block, we add a UCD layer, which consists of a convolutional layer with 1×1 kernel size and a downsampling layer which a sampling ratio is 0.5. The UCD layer allows us to speed up the network without reducing the amount of information in the input but maintaining the ratio between the information, and the size of the input is reduced to half of the original size after the UCD layer. Afterwards, the feature extraction is performed by a network layer composed of different numbers of SSN blocks, while maintaining the same resolution of the input. Due to the existence of the self-attention mechanism, SSN can capture the correlation of different local information features, so as to achieve mutual dependence between local information and global information. Finally, the results are output through an average pooling layer and a projection layer as well as a classifier layer. 

Through the historical images of driving trips, we can obtain information such as position, yaw and environment. SSN is similar to a typical vision network and can adjust the stride size of the middle layer to obtain feature maps of different sizes according to the requirements, which can be applied to downstream tasks with different inputs, such as trajectory prediction and image classification.

\subsection{SSN blcok}
The proposed SSN module consists of a Reinforcement Region Unit (RRU), a Fast Multi-Head Self-Attention (FMHSA) module and an Information Refinement Unit (IRU), as shown in Fig.~\ref{fig3}. We will describe these four components in the following.

Reinforcement region unit. In vision tasks, data augmentation is usually essential to improve model generalization effectively by training the augmented data. Common data augmentation methods such as flip, rotation and scaling, etc, but adding augmented data should not weaken the final performance of the model.

In other words, a good model should maintain effective operating output for similar but variant data as well so that the model has better input acceptability. However, the absolute position encoding used in the common attention was originally designed to exploit the order of the tokens, but it breaks the input acceptability because each patch adds a unique position encoding to it ~\cite{chu2021conditional}. Moreover, the concatenation between the local information obtained by the information capture module at the beginning of the model and the structural information inside the patch ~\cite{islam2020much} is ignored. In order to maintain input acceptability, the Reinforcement Region Unit (RRU) is designed to extract the local information from the input to the "SSN" module, defined as:
\begin{equation}
RRU(X)=Conv(Conv(X)).
\end{equation}

Fast Multiple Head Self-Attention (FMHSA) module consists of one convolutional layer, one linear layer and one multi-head self-attention. With this scheme, we can build up the connection between different local information which results from RRU. By this way, the collision avoidance task is able to get an outstanding result since on each frame the trajectory is composed of continuously predicted positions. Moreover, these positions are sequentially related which means autonomous driving vehicles must arrive at the first target position then they can move to the next target position. Our FMHSA module is suitable to solve this problem because it can transfer local information between areas.

The Information Refinement Unit (IRU) is used to efficiently extract the local information obtained by FMHSA, and after processing by this unit, the extracted local information is fed into the pooling and classifier layers. The original FFN proposed in ViT consists of two linear layers separated by the GELU activation~\cite{hendrycks2016gaussian}. First, expand the input dimension by 4 times, and then scale down the expanded dimension: 
\begin{equation}
FFN(X) = GELU(XW1 + b1)W2 + b2.
\end{equation}
This has the advantage of using a linear function for forward propagation before using the GELU() function, which greatly improves the efficiency of the model operation. However, this strategy leads to a certain performance sacrifice when the network is propagating fast in this region. Our design concept can be used to deal with this problem. First, we use the convolutional layer of a larger convolution kernel to obtain the characteristics of the input information with a large field of view, then use the linear function layer to conduct quickly, and finally use the convolutional layer of a small convolution kernel. The convolutional layer obtains refined information, thus taking into account both operational efficiency and model performance. The expression of the information refinement unit (IRU) can be written as
\begin{equation}
IRU(X)=Conv(L(Conv(X))),
\end{equation}
where L(X)=WX+b. After designing the above three unit modules, the SSN block can be formulated as: 
\begin{equation}
\begin{split}
    A&=RRU(X) \\ B&=FMHSA(A) \\ C&=IRU(B)+B
\end{split}
\end{equation}
In the experiment part, we will prove the efficiency of SSN network.
\section{Experiment}


In this section, we investigate the effectiveness of the SSN architecture by conducting experiments on an autonomous driving obstacle avoidance task based on a driving map as the main input. We compare the proposed SSN with other popular models before showing in Fig.~\ref{fig2}, and then compare the experimental results to draw an analytical conclusion. We defined three different types of collisions which are front collision, rear collision and side collision. These situations are caused by different unsuitable physic parameters and in Fig.~\ref{fig1}.

\subsection{Dataset and description}

We use l5kit dataset~\cite{houston2021one} as our data source which contains over 1,000 hours of data. This was collected by a fleet of 20 autonomous vehicles along a fixed route in Palo Alto, California, over a four-month period. It consists of 170,000 scenes, where each scene is 25 seconds long and captures the perception output of the self-driving system, which encodes the precise positions and motions of nearby vehicles, cyclists, and pedestrians over time. On top of this, the dataset contains a high-definition semantic map with 15,242 labeled elements and a high-definition aerial view over the area.

\subsection{Data preprocessing}

The data mainly includes the following main concepts: Scenes, Frames, and Agents. A scene is identified by the host (i.e. which car was used to collect it) and a start and end time. It consists of multiple frames (=snapshots at discretized time intervals). The scene datatype stores reference to its corresponding frames in terms of the start and end index within the frames array (described below). The frames in between these indices all correspond to the scene (including the start index, excluding the end index.

A frame captures all information that was observed at a time. This includes the timestamp, which the frame describes; data about the ego vehicle itself such as rotation and position; a reference to the other agents (vehicles, cyclists and pedestrians) that were captured by the ego’s sensors; a reference to all traffic light faces (see below) for all visible lanes. An agent is an observation by the AV of some other detected object. Each entry describes the object in terms of its attributes such as position and velocity, and gives the agent a tracking number to track it over multiple frames (but only within the same scene!) and its most probable label. 

The input of this dataset is images of Ego car, which is one of properties of Ego Dataset. And the output of our model are position and yaw which are properties of EgoDataset as well. By this way, we can simulate vehicles’ driving as human driving actions. During human driving process, drivers control accelerator and driving wheels to move vehicles, accelerator is used for velocity and driving wheel for yaw. The output of our model is also velocity and yaw. Thus, we use this method to simulate the trajectories of vehicles so that we can change velocity and yaw to avoid collisions during driving.

\subsection{Result}

The tables of test results which are processed by four
different network structures are shown in Tab.~\ref{tab1}. Compared with other transformer-based and convolution-based
counterparts, our model achieved better accuracy and
faster processing speed. In particular, our model achieves
2.6 times on front collision which is 13.6 times less than
RepVGG, 5.8 times less than ViT, and 12.6 times less
than ResNet50, indicating the benefit of SSN block for
capturing both local and global information. We can see
that SSN consistently outperforms other models by a large
margin.

\begin{table}[ht]
    \caption{Different collision times of four models per 10000 frames.}
    \renewcommand\arraystretch{1.5}
    \centering
    \begin{tabular}{@{\hspace{0pt}}m{3cm}<{\centering}@{\hspace{0pt}}@{\hspace{0pt}}m{1.9cm}<{\centering}@{\hspace{0pt}}@{\hspace{0pt}}m{1.9cm}<{\centering}@{\hspace{0pt}}@{\hspace{0pt}}m{1.9cm}<{\centering}@{\hspace{0pt}}}
         \hline
         \hline
        Method & Front
 & Side
 & Rear \\ 
         \hline
         L5Kit (ResNet-50)~\cite{houston2021one} &15.2 &20.7 &8.3 \\
        RepVGG~\cite{ding2021repvgg} &16.2 &11.7 &10.6 \\
        Vit~\cite{dosovitskiy2020image} &8.4 &\textbf{7.7} &9.2 \\
        SSN (Ours) &\textbf{2.6} &13.3 &\textbf{3.5} \\
         \hline
         \hline
    \end{tabular}
    \label{tab1}
\end{table}

\section{Conclusion}
This paper proposes a novel hybrid architecture named SSN for vision-based autonomous driving tasks and other vision tasks. The designed SSN architectures take advantages of both CNNs and self-attention to capture local and global information, improving the ability of the sequentially related inputs. Extensive experiments on lykit dataset demonstrate the effectiveness and superiority of the proposed SSN architecture.
\bibliographystyle{IEEEtran}
\bibliography{IEEEfull, ref}

\end{document}